\RequirePackage{fix-cm}

\documentclass[twocolumn]{svjour3}
\smartqed

\usepackage{microtype}
\usepackage{times}
\usepackage[authoryear,sort&compress]{natbib}
\usepackage{amsmath,amsfonts}
\usepackage{algorithmic}
\usepackage{algorithm}
\usepackage{array}
\usepackage[caption=false,font=footnotesize]{subfig}
\usepackage{textcomp}
\usepackage{stfloats}
\usepackage{url}
\usepackage{verbatim}
\usepackage{graphicx}
\usepackage{multirow}
\usepackage{bbding}
\usepackage{booktabs}
\usepackage{colortbl}
\usepackage[table]{xcolor} 
\usepackage{makecell}
\usepackage[pagebackref=false,breaklinks=true,colorlinks,bookmarks=false,urlcolor=magenta, citecolor=blue]{hyperref}

\newcolumntype{L}[1]{>{\raggedright\let\newline\\\arraybackslash\hspace{0pt}}m{#1}}
\newcolumntype{C}[1]{>{\centering\let\newline\\\arraybackslash\hspace{0pt}}m{#1}}
\newcolumntype{R}[1]{>{\raggedleft\let\newline\\\arraybackslash\hspace{0pt}}m{#1}}

\journalname{IJCV}
\makeatletter
\renewcommand{\makeheadbox}{}
\makeatother
\date{}
\begin{document}

\makeatother
\title{CineWeaver: Training-Free Reference-Controllable Multi-Shot \\ Long Video Generation for Cinematic Storytelling}

\author{Yuyang~Huang$^{1}$ \and Yabo~Chen$^{3}$ \and Wenrui~Dai$^{2}$ \and Ziyang~Zheng$^{2}$ \and Haibin~Huang$^{3}$ \and Chi~Zhang$^{3}$ \and Junni~Zou$^{2}$ \and Hongkai~Xiong$^{2}$ \and Xuelong~Li$^{3}$}

\authorrunning{Yuyang Huang et al.}

\institute{$^1$Department of Computer Science and Engineering, Shanghai Jiao Tong University, Shanghai 200240, China. $^2$Department of Electronic Engineering, Shanghai Jiao Tong University, Shanghai 200240, China. $^3$Institute of Artificial Intelligence (TeleAI), China Telecom, China.\\
\email{huangyuyang@sjtu.edu.cn; chenyb44@chinatelecom.cn; daiwenrui@sjtu.edu.cn; zhengziyang@sjtu.edu.cn; huanghb28@chinatelecom.cn; zhangc120@chinatelecom.cn; zoujunni@sjtu.edu.cn; xionghongkai@sjtu.edu.cn; xuelong\_li@ieee.org}
}

\maketitle

\begin{abstract}
Cinematic video generation is challenging for text-to-video diffusion models due to concurrent requirements on multi-shot generation, fine-grained controllability over characters and scenes, and long-form generation across extended temporal horizons. Existing methods rely on customization and retraining to separately address specific requirements, and cannot simultaneously fulfill all the requirements with a unified framework. In this paper, we shed light on the training-free paradigm with the key insight that the difficulty of multi-shot generation arises from a structural bias toward temporal continuity in pretrained video diffusion models, and consequently, propose a unified framework named CineWeaver to achieve reference-controllable multi-shot long-video generation without retraining. We manipulate positional encoding and attention patterns to break temporal continuity during inference to enable clear shot transitions using pretrained video diffusion models. Furthermore, we extend the proposed framework with a shot-routed reference conditioning mechanism for per-shot fine-grained controllability, and develop an anchor memory mechanism to allow long-form generation with consistent global appearance cues. To our best knowledge, CineWeaver is the first unified framework to simultaneously enable \textbf{long-form}, \textbf{reference-controllable}, and \textbf{multi-shot} video generation in a training-free fashion. Experimental results demonstrate that CineWeaver produces high-quality cinematic videos of long durations with consistent identities, stable global appearance, and clear shot transitions. The project page is available at: \url{https://cineweaver.github.io}.
\end{abstract}


\begin{figure*}[!t]
\renewcommand{\baselinestretch}{1.0} 
\centering
\includegraphics[width=\linewidth]{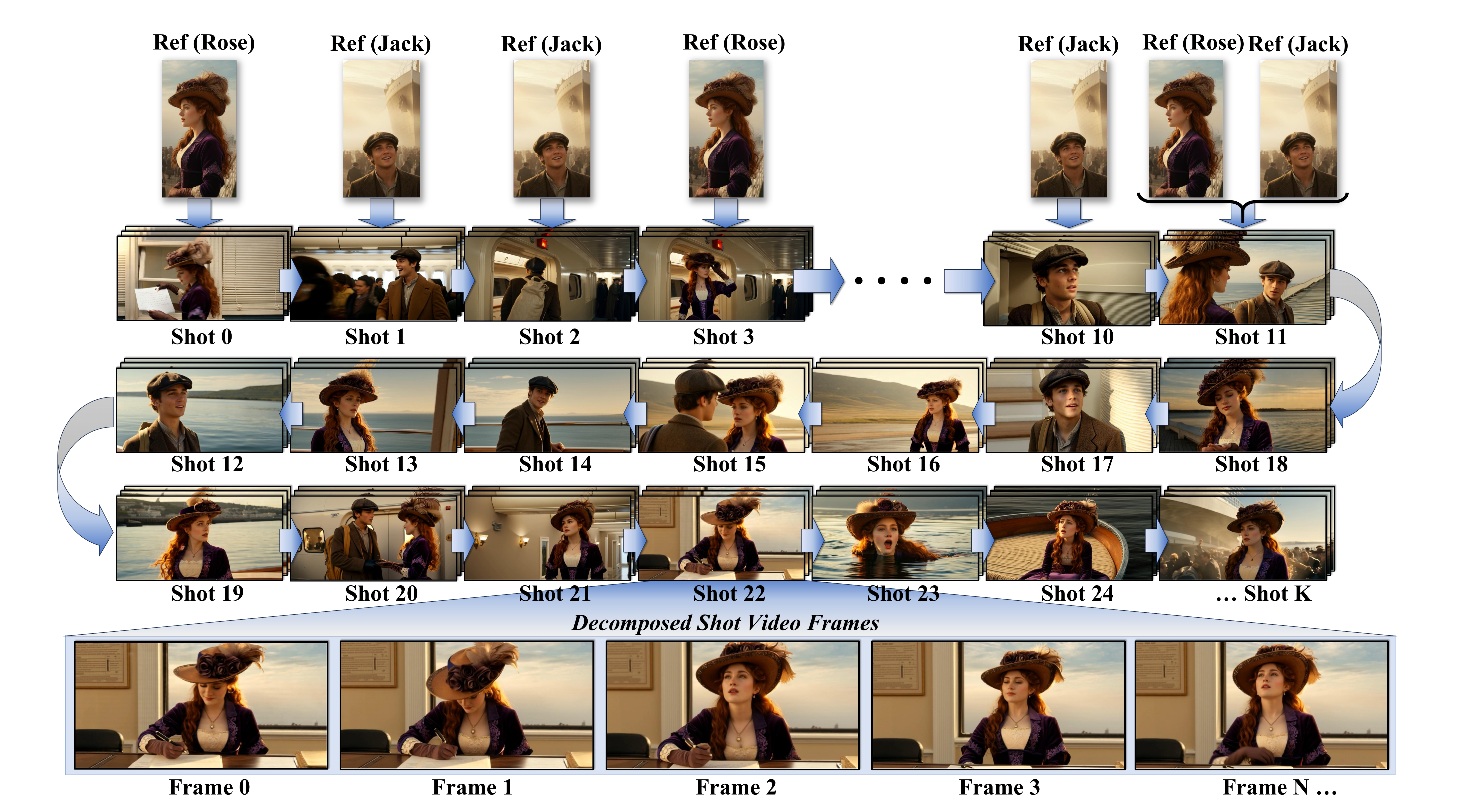}  
\caption{CineWeaver generates multi-shot \textit{Titanic}-style video conditioned on reference images of Jack and Rose and shot-level descriptions.}\label{fig:teaser}
\end{figure*}

\section{Introduction}
Video diffusion models have recently demonstrated impressive progress in synthesizing short video clips from natural language descriptions~\citep{cogvideox,wan,hunyuanvideo,longcat,skyreelsa2,lumiere,latte,packing,huang2025zero,chen2025teleworld,benchmark,chen2025tuning}. 
Beyond short-form visual synthesis, cinematic video generation represents a more challenging task for multimedia content creation that expects to transform textual scripts, visual references, and shot-level intentions into a coherent long-form narrative video. It is central to emerging applications such as film pre-visualization, digital storytelling, advertising content creation, and interactive multimedia production~\citep{aiflow,aiflow_edge,wang2026directing,chen2026full,zhang2026telephysics,zhang2026symphomotion,chen2024liftimage3d,chen2024cascade,zhou20253dgabsplat}.

Nevertheless, cinematic video generation remains challenging to existing methods due to three requirements of film production. First, cinematic storytelling inherently relies on \textbf{multi-shot generation}, where a narrative is unfolded through a sequence of shots with different viewpoints, scales, and camera movements. Second, effective filmmaking requires \textbf{fine-grained controllability} to precisely manipulate the time, location, and development of various semantic elements, including characters, objects, and scenes. Third, cinematic content typically involves \textbf{long-form generation} that requires coherence across extended temporal horizons. Existing pretrained video diffusion models are primarily designed for single-shot clip generation and lack of the ability to naturally support multi-shot composition~\citep{cogvideox,wan,hunyuanvideo,lumiere,latte}. Moreover, there is no native mechanism for the single-shot setting to simultaneously achieve fine-grained appearance control and scalable long-form synthesis due to limited conditioning flexibility and temporal consistency issues. Recent attempts usually address these challenges separately under different temporal modeling assumptions and specialized architectural designs or training strategies~\citep{cinetrans,multishotmaster,echoshot,holocine,phantom,vace,skyreelsa2,longcat,packing,selfforcing,rollingforcing,causvid,streamingt2v,lol,framepack}. It is difficult to achieve a unified framework by directly composing these methods. 



A fundamental reason for this limitation is that existing methods resort to explicit learning for multi-shot generation via architecture customization and retraining. They introduce specialized positional encodings, transition-aware attention modules, or dedicated multi-shot architectures, and fine-tune the underlying video diffusion models on large-scale multi-shot datasets to learn shot transition patterns \citep{cinetrans,multishotmaster,echoshot,holocine}. Specific temporal pattern for shot discontinuity are modeled and embedded into learned parameters by different model architectures and training strategies. Moreover, reference control and long-form generation also rely on temporal representations and attention interactions to propagate visual identity or long-range information. Consequently, existing multi-shot approaches could be interfered by simply combining independently learned solutions to the three requirements with shared temporal mechanisms. They are specialized for modeling shot transitions and can be hardly extended to a unified framework with reference controllability and long-form generation.


Contrary to existing approaches, we argue that pretrained text-to-video diffusion models already possess the capacity to compose visually distinct scenes by learning rich scene-level priors, semantic variation, and appearance synthesis from large-scale video datasets. The main obstacle is the structural bias toward temporal continuity imposed by the native generation process, since adjacent frames are encouraged to evolve as a single continuous sequence with shot temporal positional encoding, global self-attention, prompt mixing, and VAE temporal caching in existing approaches. This fact suggests the potential of training-free multi-shot generation by addressing the structural bias during inference.

Based on this insight, we propose CineWeaver, a training-free unified framework for reference-controllable multi-shot long video generation that simultaneously supports multi-shot generation, reference controllability, and long-form synthesis within a pretrained video diffusion model without retraining. We first enable multi-shot composition by explicitly breaking temporal continuity during inference through RoPE manipulation, masked self-attention, shot-wise cross-attention and FFN, and isolated VAE decoding. Under such a training-free multi-shot structure, reference tokens are routed to designated shots for fine-grained controllability, and anchor memory propagates global appearance cues across independently generated segments to maintain long-form consistency. 
As demonstrated in Fig.~\ref{fig:teaser}, CineWeaver enables cinematic video synthesis with clear shot transitions, fine-grained reference control, and consistent global appearance over long sequences. Extensive experiments show that CineWeaver achieves strong shot transition quality, reference consistency, and long-form coherence across diverse cinematic video generation scenarios.

The contributions of this paper are summarized as follows.
\begin{itemize}
\item We shed light on the training-free paradigm as a unified framework for multi-shot generation based on pretrained video diffusion models. We successfully achieve multi-shot generation via training-free inference-time operations, including gap-frame RoPE manipulation, start-frame-aware masked self-attention, shot-wise conditioning, and isolated VAE decoding.
\item We present a reference-controllable extension for multi-shot generation with shot-routed self-attention and cross-attention modules. The extended multi-shot generation framework supports reference-controllable generation with cross-shot identity consistency and enables explicit per-shot control on characters, objects, and scenes. 
\item We design an anchor memory mechanism that provides a simple yet effective inference path with time-invariant anchor token caching and anchor-aware attention mask toward cinematic-level controllable narrative video synthesis. The proposed mechanism can scale to arbitrarily long videos with ensured global visual consistency and precise reference control across extended sequences. 
\end{itemize}

The rest of this paper is organized as follows. In Section \ref{sec:related}, we review related work on multi-shot, reference-guided, and long video generation. Section \ref{sec:method} elaborates the proposed training-free CineWeaver for reference-controllable multi-shot long video generation, and Section \ref{sec:exp} presents experimental results for cinematic and animated multi-shot video generation. Finally, we draw conclusions in Section \ref{sec:con}.

\section{Related Work}\label{sec:related}
\subsection{Multi-shot Video Generation}
Multi-shot video generation synthesizes videos composed of multiple semantically distinct shots, with each shot corresponding to a coherent scene segment with a clear temporal boundary. Existing methods enable multi-shot generation with video diffusion models through three main mechanisms. First, \textit{redesigning temporal positional encoding}~\citep{echoshot,multishotmaster} modifies the temporal representation to explicitly encode shot identity. These methods assign segment-aware positional embeddings or introduce discontinuous temporal indices to occupy separate regions in the temporal embedding space with frames from different shots. Second, \textit{specialized attention mechanism}~\citep{holocine} redesigns temporal attention modules to explicitly model intra-shot and inter-shot interactions with hierarchical attention patterns that separate local (within-shot) and global (cross-shot) dependencies. Third, \textit{explicit transition-aware conditioning} \citep{cinetrans} injects additional control signals such as learned transition tokens or boundary indicators into the diffusion process to guide shot transitions and explicitly trigger scene changes.

These methods commonly require dedicated multi-shot training data and specialized model architecture and training procedure to learn to enable multi-shot generation. 
This tight coupling restricts the learned model to the specific capability of shot cutting, and architectural modifications and retraining are required to incorporate reference-guided generation or long-form modeling in addition to shot cutting. 
In fact, we find that multi-shot generation can be realized by manipulating temporal continuity during inference instead of retraining. Shot composition can be decoupled from model training and architectural design by introducing shot boundaries to explicitly interrupt the generation process along the temporal dimension. 

\subsection{Reference-Guided Video Generation}
Reference-guided video generation focuses on improving controllability by conditioning the generation process on external visual references. Existing approaches introduce reference images or identity tokens into video diffusion models through mechanisms such as cross-attention conditioning~\citep{vace,customvideox,contextanyone}, visual token injection~\citep{phantom,skyreelsa2,magicmirror}, reference encoders \citep{kaleido,saber}, and subject customization~\citep{customvideo,videodreamer}. These methods effectively enforce appearance consistency by propagating reference information throughout the entire denoising trajectory, and are well-suited for single-shot video generation. 

These approaches apply reference information at a \textit{global clip level} to govern the entire temporal sequence with a single conditioning signal. This design implicitly assumes temporal continuity within the video, and does not align with multi-shot generation that requires distinct semantic contexts and potentially different visual roles. In essence, existing approaches are incompatible with shot-level temporal decomposition by entangling reference conditioning with global temporal coherence. 
On the contrary, we decouple reference conditioning from global temporal continuity with shot-level reference routing during inference to apply different references to different shots in a unified generation process.


\subsection{Long Video Generation}
Long video generation with coherent appearance and motion over extended temporal horizons remains a fundamental challenge for diffusion-based video models. Existing methods primarily extend video length by preserving temporal continuity under a single-shot assumption. Representative methods include overlapping window generation~\citep{streamingt2v,freelong,longdiff}, latent reuse~\citep{longcat}, autoregressive generation~\citep{causvid,selfforcing,rollingforcing,diffusionforcing,peng2026towards,xiang2025macro}, temporal modeling modules~\citep{lstd}, and memory-based mechanisms \citep{longlive,malt,memorypack,framepack,packing}. 
Although these methods can maintain long-range consistency, they fundamentally operate under the assumption of a single continuous scene. This leads to a structural mismatch with multi-shot generation, where the temporal domain is explicitly segmented to correspond to distinct semantic units. Consequently, existing long-video methods are not naturally compatible with shot-level decomposition or reference routing by enforcing continuous temporal dependency rather than based on segmented temporal structure. In contrast, we explicitly introduce temporal discontinuities during generation to enable a unified framework that supports long-range coherence and shot-level composition without architectural changes or retraining.


\section{Proposed Method}\label{sec:method}

\subsection{Problem Definition and Motivation}
\label{subsec:preliminaries}

In this paper, we focus on reference-controllable multi-shot long video generation. Formally, given a sequence of shot-level text prompts 
$\mathcal{P} = \{p_i\}_{i=1}^{N_K}$, 
corresponding shot lengths 
$\mathcal{L} = \{l_i\}_{i=1}^{N_K}$, 
and optional shot-wise reference sets 
$\mathcal{R} = \{\mathcal{R}_i\}_{i=1}^{N_K}$, we aim to generate a composed multi-shot video $\mathbf{V} = [\mathbf{v}_1; \mathbf{v}_2; \cdots; \mathbf{v}_{N_K}]$, where each shot $\mathbf{v}_i \in \mathbb{R}^{l_i \times H \times W \times 3}$. The generated video should simultaneously satisfy four constraints: i) Each shot should semantically align with its corresponding prompt $p_i$,
ii) Each shot should preserve the visual attributes from its reference set $\mathcal{R}_i$ if provided,
iii) Adjacent shots should exhibit clear transitions,
iv) Temporal coherence should be maintained within each shot, while Global coherence in identity, style, and layout is further required across the full video sequence.

Among the three constraints, multi-shot generation is the most fundamental and difficult to satisfy under existing video diffusion frameworks. Existing methods usually introduce specialized architectures or multi-shot video datasets and require retraining to explicitly model shot transitions. They become specialized for modeling shot transitions, and the generation process is specialized toward temporal shot cutting. This introduces conflicts with other generation objectives that rely on different temporal modeling behaviors, such as global identity propagation in reference-guided generation and long-range dependency modeling in long-form synthesis. We argue that multi-shot generation does not necessarily require learning, but can be achieved as a structural operation on the inference process by explicitly interrupting temporal continuity. Therefore, shot composition is decoupled from model parameter learning to eliminate the conflict in satisfying all the constraints.

To this end, we first enable training-free multi-shot T2V generation by allowing a pretrained T2V model to perform shot composition during inference without retraining or modifying its learned priors in Section~\ref{subsec:multishot_t2v}. Building upon this multi-shot structure, we design a reference routing mechanism for shot-wise reference control in Section~\ref{subsec:reference_multishot}, and an anchor memory mechanism for long-form generation with cross-shot long-range consistency in Section~\ref{subsec:long_video}.

\begin{figure*}[!t]
\renewcommand{\baselinestretch}{1.0}
\setlength{\abovecaptionskip}{0pt}
\centering
\includegraphics[width=\linewidth]{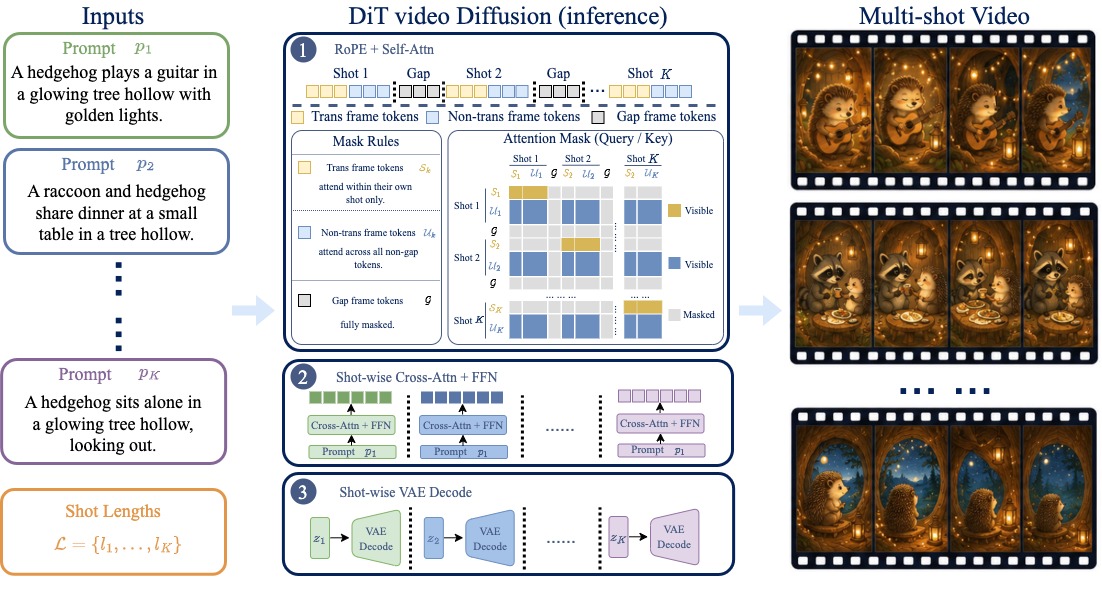}
\caption{We insert \textit{gap frames} to weaken RoPE-induced coupling, and apply a temporarily isolated self-attention mask with \textit{transition frames} for a clear cut. We adopt shot-wise cross-attention and FFN to avoid inter-shot prompt interference, and shot-wise VAE decoding to avoid cache-induced visual leakage.}\label{fig:pipeline}
\end{figure*}

\subsection{Training-Free Multi-Shot T2V Generation}
\label{subsec:multishot_t2v}
Pretrained video diffusion models tend to generate temporally continuous single-shot clips due to structural continuity biases in the native inference pipeline, including RoPE-based temporal proximity, global self-attention across adjacent frames, shared prompt conditioning, and temporal caching in VAE decoding. Therefore, we selectively relax these continuity biases during inference to synthesize distinct scenes for multi-shot generation without introducing trainable modules. As shown in \figurename~\ref{fig:pipeline}, our approach consists of three key designs: 1) Shot-aware RoPE and masked self-attention, 2) Shot-wise cross-attention and FFN, and 3) Shot-wise VAE decoding.

\subsubsection{Shot-aware RoPE and Masked Self-Attention}\label{gap_mask}
We consider the two primary mechanisms of rotary positional encoding (RoPE) and self-attention that induce temporal continuity in video diffusion models. Under RoPE, attention scores between temporally adjacent frames are typically higher due to smaller relative positional offsets, leading to stronger inter-frame coupling and resulting in temporally continuous visual content. Self-attention enables global information exchange across frames by coupling their representations and reducing temporal independence to promote temporally consistent generation across adjacent frames. We develop shot-aware RoPE and masked self-attention to disrupt the temporal continuity of the generation process and enable shot-level separation.

\noindent\textbf{Shot-aware RoPE with gap frames.} In standard DiT-based video diffusion models, RoPE~\citep{rope} directly injects relative temporal information into the attention computation, and is not aware of shot cuts due to higher attention scores for temporally adjacent frames. 
Specifically, the attention score between the $i$-th query token $\mathbf{q}_i$ and the $j$-th key token $\mathbf{k}_j$ is:
\begin{equation}\label{eq:rope_attn}
\text{Self-Attn}(i, j) \propto \mathbf{q}_i^\top R_{j-i} \mathbf{k}_j,
\end{equation}
where $R_{j-i}$ denotes the rotary transformation defined by the relative temporal offset $\Delta d = j - i$ between $\mathbf{q}_i$ and $\mathbf{k}_j$. A smaller $|\Delta d|$ generally leads to a higher $\text{Self-Attn}(i, j)$. To mitigate this, unlike customized positional embeddings~\citep{echoshot,multishotmaster}, we introduce $N_G$ \emph{gap frames (G-frames)} as a fixed offset between every two shots for enlarging $|\Delta d|$ to $|\Delta d|+|\mathcal{G}|$ to preserve the RoPE formulation and avoid retraining, where $\mathcal{G}$ is the index set of tokens from all G-frames. Tokens from G-frame are excluded from all valid attention interactions and do not participate in self-attention.

\noindent\textbf{Masked self-attention with transition frames.} Moreover, we introduce $N_S$ \emph{transition frames} at the beginning of each shot and employ masked self-attention to explicitly prevent early cross-shot token interaction, thereby enabling the formation of clear shot boundaries. Given the index set of video tokens $\mathcal{V}_k$ in the $k$-th shot, let $\mathcal{S}_k \subset \mathcal{V}_k$ denote the index set of tokens from the $N_S$ transition frames and $\mathcal{U}_k = \mathcal{V}_k \setminus \mathcal{S}_k$ denote the index set of remaining tokens. 
We introduce an additive mask matrix $\mathbf{M}$ to the attention logits, where the element $M_{i,j}$ for the $i$-th query token and $j$-th key token is 
\begin{equation}\label{eq:attn_mask}
M_{i,j} =
\begin{cases}
0, & \text{if } i \in \mathcal{S}_k \text{ and } j \in \mathcal{V}_k \\
0, & \text{if } i \in \mathcal{U}_k \text{ and } j \in \bigcup_m \mathcal{V}_m \\
-\infty, & \text{otherwise}
\end{cases}.
\end{equation}
\figurename~\ref{fig:attentionmask} illustrates the masked self-attention. Tokens from transition frames are restricted to attend only to video tokens within their own shot, enforcing boundary isolation at the early stage of each shot. The remaining tokens are visible to all video tokens except for G-frames, enabling cross-shot interaction to share global context and enhance cross-shot consistency.

\subsubsection{Shot-wise Cross-Attention and FFN}
Shot-wise cross-attention strategy is naturally adopted for the multi-shot setting with shot-specific prompts. The cross-attention computation is strictly applied to the query video tokens within each shot. For the $k$-th shot, given the quey video tokens $\mathbf{Q}_k$,
\begin{equation}\label{eq:shot_cross_attn}
\text{Cross-Attn}(\mathbf{Q}_k, \mathbf{K}, \mathbf{V}) \longrightarrow \text{Cross-Attn}(\mathbf{Q}_k, \mathbf{K}_k, \mathbf{V}_k),
\end{equation}
where $\mathbf{K}_k$ and $\mathbf{V}_k$ are the key and value tokens from the $k$-th prompt $p_k$. Note that G-frames are removed here, since cross-attention does not involve RoPE. The FFN transformation is similarly realized for each shot independently to avoid mixing residual features across shot boundaries. 

\subsubsection{Shot-wise VAE Decoding} 
In DiT-based video diffusion models, the VAE usually suffers from cache-induced leakage due to temporal caching mechanisms for improved efficiency. When the entire latent sequence is decoded jointly, cached states from a preceding shot could implicitly influence subsequent shots and cause unintended visual leakage across shots. 

Shot-wise decoding is designed to partition the latents $\mathbf{z}$ from the final denoising step into $N_K$ subsequences $\mathbf{z}_1, \cdots, \mathbf{z}_{N_K}$ of predefined shot lengths $\mathcal{L}=\{l_k\}_{k=1}^{N_K}$. 
For any $k = 1, \dots, N_K$, $\mathbf{z}_k$ corresponds to the $k$-th shot and is decoded independently by $\mathbf{v}_k = \mathrm{VAE}(\mathbf{z}_k)$ with the VAE decoding state reset to ensure that no temporal cache is shared across shots. 
Moreover, the self-attention of the $N_S$ \emph{transition frames} at the beginning of each shot is restricted to the tokens within the same shot, preventing early cross-shot interaction and establishing clear shot boundaries.
Due to their constrained interaction during denoising, the first transition frame of each shot may contain boundary artifacts. 
To improve perceptual quality, we apply a lightweight boundary refinement strategy at the decoding stage. 
Specifically, for the $N_S$ transition latent frames of each shot, we replace the first latent frame with the second one before VAE decoding, and denote the replaced frame as a \emph{dummy latent frame}. 
After decoding, this dummy latent frame produces a dummy pixel frame, which is directly discarded from the final output. 
Finally, the remaining decoded segments are concatenated to form the composed video $\mathbf{V} = [\mathbf{v}_1; \cdots; \mathbf{v}_{N_K}]$.

\subsection{Reference-Controllable Multi-Shot Generation}
\label{subsec:reference_multishot}

We extend the training-free multi-shot T2V to achieve shot-wise reference control by routing visual reference tokens according to the shot structure, rather than redesigning or retraining the reference-guided backbone. Existing methods such as Phantom~\citep{phantom} allow for controlling single-shot generation by introducing the tokens of reference images to self-attention. On the contrary, we achieve structured multi-shot reference-controllable generation to compose the cinematic video $\mathbf{V}=[\mathbf{v}_1;\cdots;\mathbf{v}_{N_K}]$ complying with a set of reference images $\mathcal{R}=\{\mathcal{R}_k\}_{k=1}^{N_K}$ for arbitrary $K$ shots (one or more reference images for each shot). Any shot $\mathbf{v}_k$ is generated by matching $\mathcal{R}_k$ in appearance (\emph{e.g.}, identity, texture, layout) and following the motion described by the prompt $p_k$. To this end, we devise reference tokens $\mathbf{h}^{\text{ref}}$ from $\mathcal{R}$ invariant to denoising steps and design shot-wise self-attention and cross-attention mechanisms for transition and non-transition video tokens in each shot to interact with corresponding reference tokens.

Reference images are encoded into clean latents $\mathbf{z}^{\text{ref}}$, patchified into reference tokens $\mathbf{h}^{\text{ref}}$, and concatenated with video tokens $\mathbf{h}_t^{\text{vid}}$ derived from noisy video latents $\mathbf{z}_t^{\text{vid}}$ at each denoising step.
\begin{equation}
\mathbf{H}_t = [\mathbf{h}_t^{\text{vid}}; \mathbf{h}^{\text{ref}}].
\end{equation}
Self-attention allows $\mathbf{h}_t^{\text{vid}}$ to attend to the noise-free, time-invariant $\mathbf{h}^{\text{ref}}$ for stable visual anchors to preserve reference appearance. We apply this clean-to-noisy conditioning per shot, injecting $\mathcal{R}_k$ only into its corresponding shot tokens while preventing cross-shot semantic leakage as discussed in Section~\ref{subsec:multishot_t2v}.

\begin{figure*}[!t]
\renewcommand{\baselinestretch}{1.0} 
\setlength{\abovecaptionskip}{0pt}
\centering
\includegraphics[width=\linewidth]{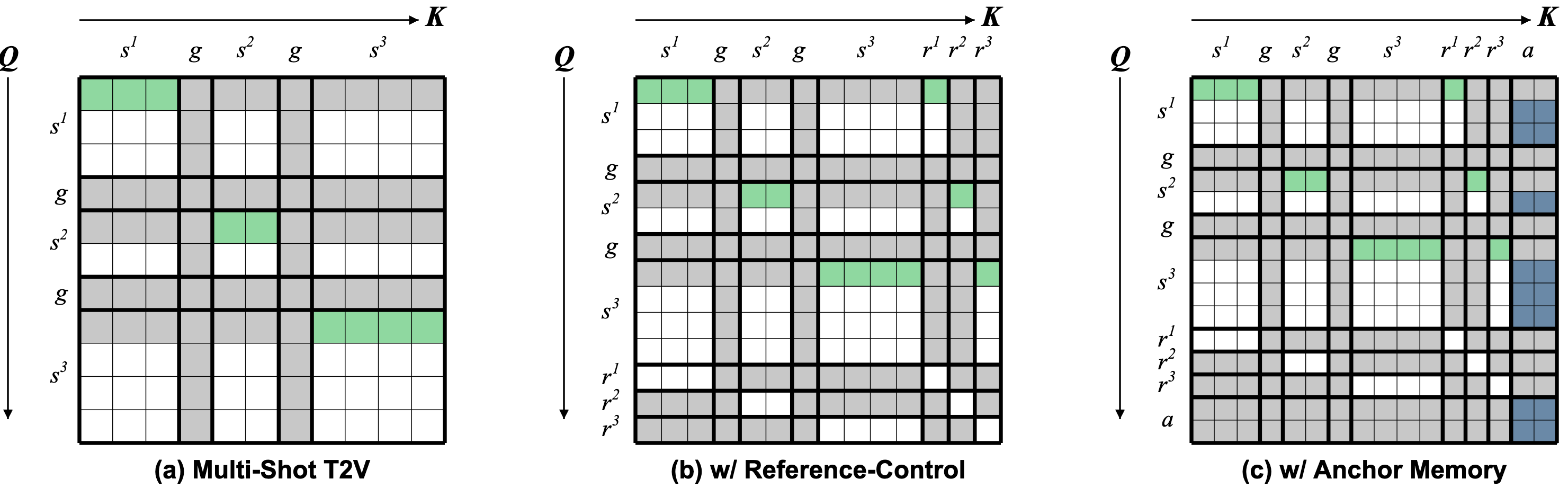}  
\caption{Attention masks for (a) Multi-Shot T2V, (b) Reference-Controllable Multi-Shot T2V, and (c) Long Video Extension via Anchor Memory. $g$, $s^i$, $r^i$, and $a$ denote the tokens of gap frames, the video tokens of the $i$-th shot, the reference tokens of the $i$-th shot, and the anchor memory, respectively. Gray cells indicate masked tokens, while other colored cells (including white) denote visible attention. Green denotes the attention of transition frames tokens, and blue indicates the attention of anchor tokens.}\label{fig:attentionmask}
\end{figure*}

\subsubsection{Shot-wise Self-Attention With Reference Tokens}
As shown in Fig.~\ref{fig:attentionmask}, we design a joint masked self-attention mechanism by extending the attention mask matrix $\mathbf{M}$ in~\eqref{eq:attn_mask}.
\begin{equation}\label{eq:joint_attn_mask}
M_{i,j} = 
\begin{cases} 
0, & \text{if } i \in \mathcal{S}_k \text{ and } j \in (\mathcal{V}_k \cup \mathcal{R}_k) \\
0, & \text{if } i \in \mathcal{U}_k \text{ and } j \in (\bigcup_m \mathcal{V}_m \cup \mathcal{R}_k) \\
0, & \text{if } i \in \mathcal{R}_k \text{ and } j \in (\mathcal{V}_k \cup \mathcal{R}_k) \\
-\infty, & \text{otherwise}
\end{cases}.
\end{equation}
Eq.~\eqref{eq:joint_attn_mask} allows tokens from transition and non-transition frames to access their reference tokens and restricts reference tokens to only interact within their own shot.

\subsubsection{Shot-wise Cross-Attention and FFN} 
We further develop the shot-wise cross-attention and FFN for video tokens to incorporate the reference tokens. Similar to the shot-wise strategy elaborated in Section~\ref{subsec:multishot_t2v}, we apply the same localized processing to prevent semantic mismatch for newly introduced reference tokens. 
Let $\mathbf{Q}_k^{\text{ref}}$ denote the query tokens derived from $\mathcal{R}_k$ of the $k$-th shot. During cross-attention, $\mathbf{Q}_k^{\text{ref}}$ exclusively attends to the key tokens $\mathbf{K}_k$ and value tokens $\mathbf{V}_k$ derived from its corresponding shot-specific prompt $p_k$. Subsequently, the FFN transformations for these reference tokens are computed independently within each shot to prevent leakage across different shots.

\subsection{Long Video Extension via Anchor Memory}
\label{subsec:long_video}

Long-form synthesis of cinematic videos containing dozens of shots exceeds the temporal budget of bidirectional DiT-based diffusion models, since performing full self-attention over the entire video sequence is computationally prohibitive. Therefore, instead of directly generating the entire long video in a single diffusion trajectory, we decompose it into multiple shorter multi-shot segments. Each segment contains several consecutive shots and is generated independently using the reference-controllable multi-shot framework described in Section~\ref{subsec:reference_multishot}. The final long video is obtained by temporally concatenating all generated segments.

However, independent segment generation introduces a new challenge. Although reference conditioning preserves shot-level appearance attributes (\emph{e.g}., identity, and objects), global visual attributes (\emph{e.g.}, lighting, color grading, and overall style) may remarkably drift across independently generated segments. This occurs because such global consistency is implicitly maintained by shared self-attention interactions within a single diffusion trajectory, which no longer exists when generation is divided into multiple independent segments.

To address this issue, we introduce Anchor Memory, which transfers global appearance cues from previously generated segments to subsequent segments. Anchor Memory extends the implicit global consistency mechanism within a single diffusion trajectory to independently generated segments, enabling scalable long-form generation while preserving cross-segment visual coherence. Specifically, we first select anchor frames from previously generated segments and encode them into time-invariant anchor tokens. These tokens are then incorporated into the generation of subsequent segments through anchor-aware self-attention, allowing non-transition video tokens to access global appearance cues while preserving shot-specific structures. Since each segment only depends on the lightweight anchor tokens rather than the entire preceding video sequence, different segments can be generated independently and parallelized across multiple GPUs.



\subsubsection{Anchor Memory Construction}
Anchor memory consists of $N_A$ \emph{anchor frames} selected from the first shot of the initial segment generated using the reference-controllable multi-shot framework in Section~\ref{subsec:reference_multishot}. 
Anchor frames are jointly encoded by the VAE encoder to produce anchor latents $\mathbf{z}^{\text{anchor}}$ that are then patchified and embedded into anchor tokens $\mathbf{h}^{\text{anchor}}$. $\mathbf{h}^{\text{anchor}}$ is invariant to diffusion time steps and is concatenated with $\mathbf{H}_t$ for the subsequent segment at timestep $t$ to obtain  $\mathbf{H}_t'$ for processing with Transformer.
\begin{equation}
\mathbf{H}_t'=[\mathbf{H}_t;\mathbf{h}^{\text{anchor}}]=[\mathbf{h}_t^{\text{vid}};\mathbf{h}^{\text{ref}};\mathbf{h}^{\text{anchor}}].
\end{equation}

\subsubsection{Anchor-Aware Self-Attention}
We introduce a dedicated attention masking strategy for anchor tokens to enable stylistic propagation without structural interference. The set of anchor tokens $\mathcal{A}$ is incorporated to extend the mask matrix $\mathbf{M}$ in~\eqref{eq:joint_attn_mask}.
\begin{equation}\label{eq:anchor_attn_mask}
\!M_{i,j}\!=\! 
\begin{cases} 
0, &\text{if } i \in \mathcal{S}_k \text{ and } j \in (\mathcal{V}_k \cup \mathcal{R}_k)\\
0, &\text{if }  i \in \mathcal{U}_k \text{ and } j \in (\bigcup_m \mathcal{V}_m \cup \mathcal{R}_k \cup \mathcal{A})\\
0, &\text{if }  i \in \mathcal{R}_k \text{ and } j \in (\mathcal{V}_k \cup \mathcal{R}_k)\\
0, &\text{if }  i \in \mathcal{A} \text{ and } j \in \mathcal{A}\\
-\infty, & \text{otherwise}
\end{cases}.
\end{equation}
\figurename~\ref{fig:attentionmask} illustrates the masked attention that balances stylistic propagation with structural integrity. Non-transition tokens $\mathcal{U}_k$ are permitted to attend to anchor tokens $\mathcal{A}$ to merge global lighting, color grading, and overarching stylistic cues, while $\mathcal{A}$ is restricted to themselves to prevent their representations from being altered by the generation context.

\begin{table*}[!t]
\renewcommand{\baselinestretch}{1.0} 
\renewcommand{\arraystretch}{1.0} 
\setlength{\tabcolsep}{0.5pt}
\centering
\caption{Quantitative comparison for multi-shot text-to-video, reference-controllable multi-shot generation, and long-form multi-shot generation.}\label{tab1:main results}
\begin{tabular}{@{}l c ccc c c cc@{}}
\toprule
& \multirow{2}*{\textbf{Text Align}$\uparrow$} & \multicolumn{3}{c}{\textbf{Inter-Shot Consistency}$\uparrow$} & \multirow{2}*{\textbf{Transition Deviation}$\downarrow$} & \multirow{2}*{\textbf{Narrative \textbf{Coherence}$\uparrow$}} & \multicolumn{2}{c}{\textbf{Reference Consistency}$\uparrow$} \\
\cmidrule(lr){3-5}\cmidrule(lr){8-9}
 & & \textbf{Semantic} & \textbf{Subject} & \textbf{Scene} & & & \textbf{Background} & \textbf{Subject} \\
\midrule
\multicolumn{9}{c}{\textit{Multi-shot Text-to-Video}} \\
\midrule
CineTrans~\citep{cinetrans}          & 0.233 & 0.767 & 0.693 & 0.777 & 13.12 & 0.089 & \XSolidBrush & \XSolidBrush \\
EchoShot~\citep{echoshot}           & 0.231 & 0.736 & 0.693 & 0.744 &  8.49 & 0.144 & \XSolidBrush & \XSolidBrush \\
HoloCine~\citep{holocine}           & \textbf{0.236} & 0.681 & 0.633 & 0.691 & 14.97 & 0.235 & \XSolidBrush & \XSolidBrush \\
MultishotMaster~\citep{multishotmaster}    & 0.225 & 0.565 & 0.408 & 0.454 &  8.79 & 0.209 & \XSolidBrush & \XSolidBrush \\
\rowcolor{gray!12}
\textbf{Proposed (w/o Ref)} & \textbf{0.236} & \textbf{0.830} & \textbf{0.694} & \textbf{0.779} & \textbf{3.86} & \textbf{0.251} & \XSolidBrush & \XSolidBrush \\
\midrule
\multicolumn{9}{c}{\textit{Reference-Controllable Multi-shot Text-to-Video}} \\
\midrule
Phantom~\citep{phantom}            & \textbf{0.264} & 0.800 & 0.785 & 0.829 &  6.22 & 0.393 & 0.673 & 0.528 \\
VACE~\citep{vace}               & \textbf{0.264} & 0.818 & 0.750 &\textbf{ 0.897} & 6.90 & 0.460 & \textbf{0.744} & 0.542 \\
\rowcolor{gray!12}
\textbf{Proposed (w/ Ref)}  & 0.259 & \textbf{0.902} & \textbf{0.808} & 0.880 & \textbf{6.09} & \textbf{0.699} & 0.632 & \textbf{0.560} \\
\midrule
\multicolumn{9}{c}{\textit{Long Reference-Controllable Multi-shot Text-to-Video}} \\
\midrule
Phantom~\citep{phantom}            & 0.224 & 0.524 & 0.395 & 0.300 &  5.50 & 0.419 & 0.519 & 0.328 \\
VACE~\citep{vace}               & \textbf{0.232} & 0.533 & 0.401 & 0.278 & 4.10 & 0.545 & \textbf{0.564} & 0.361 \\
\rowcolor{gray!12}
\textbf{Proposed (w/ Ref)}  & 0.206 & \textbf{0.556} & \textbf{0.429} & \textbf{0.356} & \textbf{2.02} & \textbf{0.672} & 0.463 & \textbf{0.476} \\
\bottomrule
\end{tabular}%
\end{table*}

\begin{figure*}[!t]
\renewcommand{\baselinestretch}{1.0} 
\setlength{\abovecaptionskip}{0pt}
\centering
\includegraphics[width=0.99\linewidth]{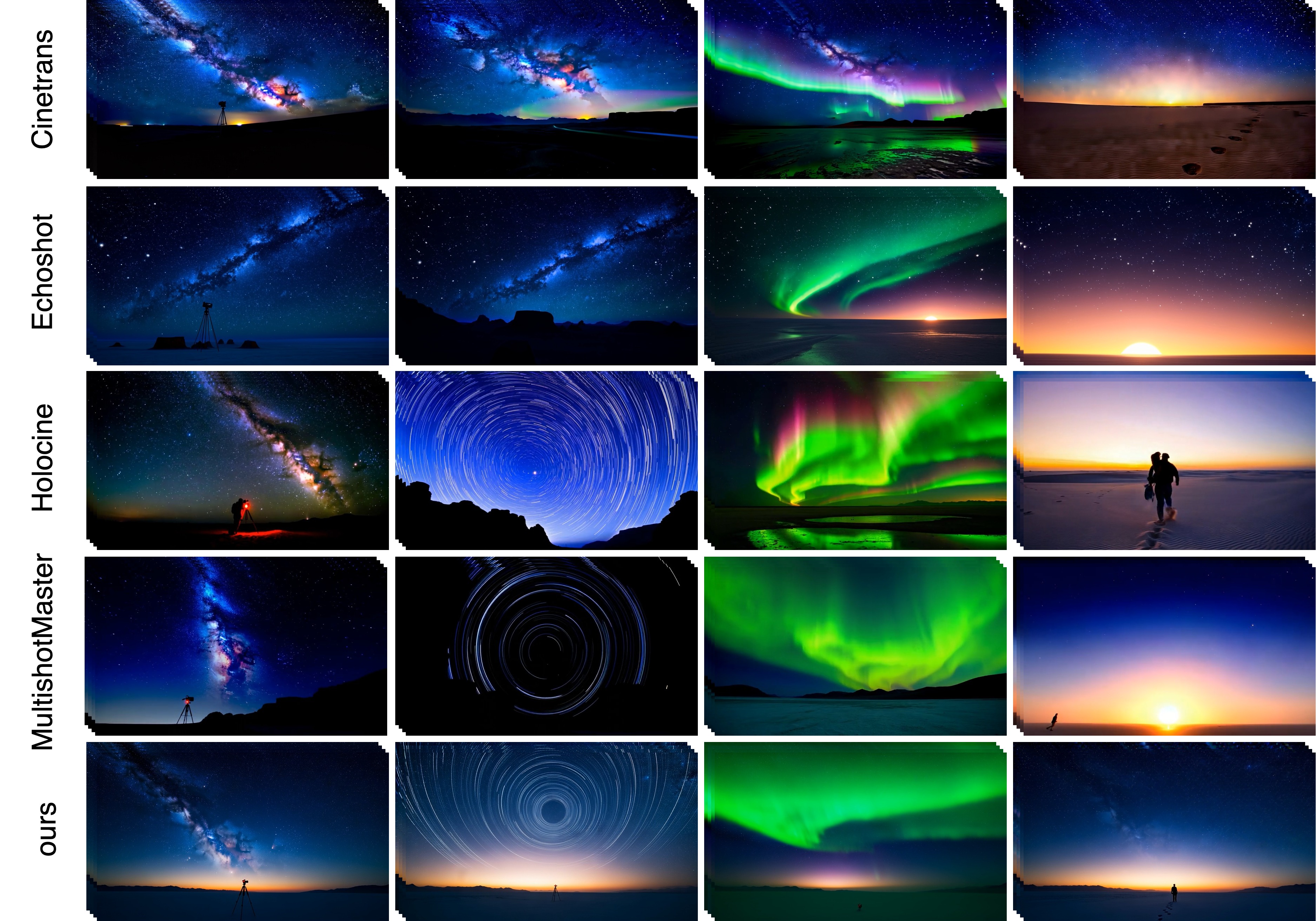}  
\caption{Qualitative Results for Multi-Shot T2V.}\label{fig:t2v-compare}
\end{figure*}

\subsubsection{Anchor Prompt Conditioning and FFN}
We further leverage the prompt embedding of the text prompt associated with the anchor frames to maintain semantic alignment. Anchor tokens perform shot-independent cross-attention with the corresponding anchor prompt embedding by $\text{Cross-Attn}$ $(\mathbf{Q}^{\text{anchor}}, \mathbf{K}^{\text{anchor}}, \mathbf{V}^{\text{anchor}})$, and do not share prompt embeddings with newly generated shots. Similarly, FFN transformations for anchor tokens are computed independently. This preserves the semantic integrity of the anchor memory while preventing cross-segment semantic leakage.



\begin{figure*}[!t]
\renewcommand{\baselinestretch}{1.0} 
\setlength{\abovecaptionskip}{0pt}
\centering
\includegraphics[width=\linewidth]{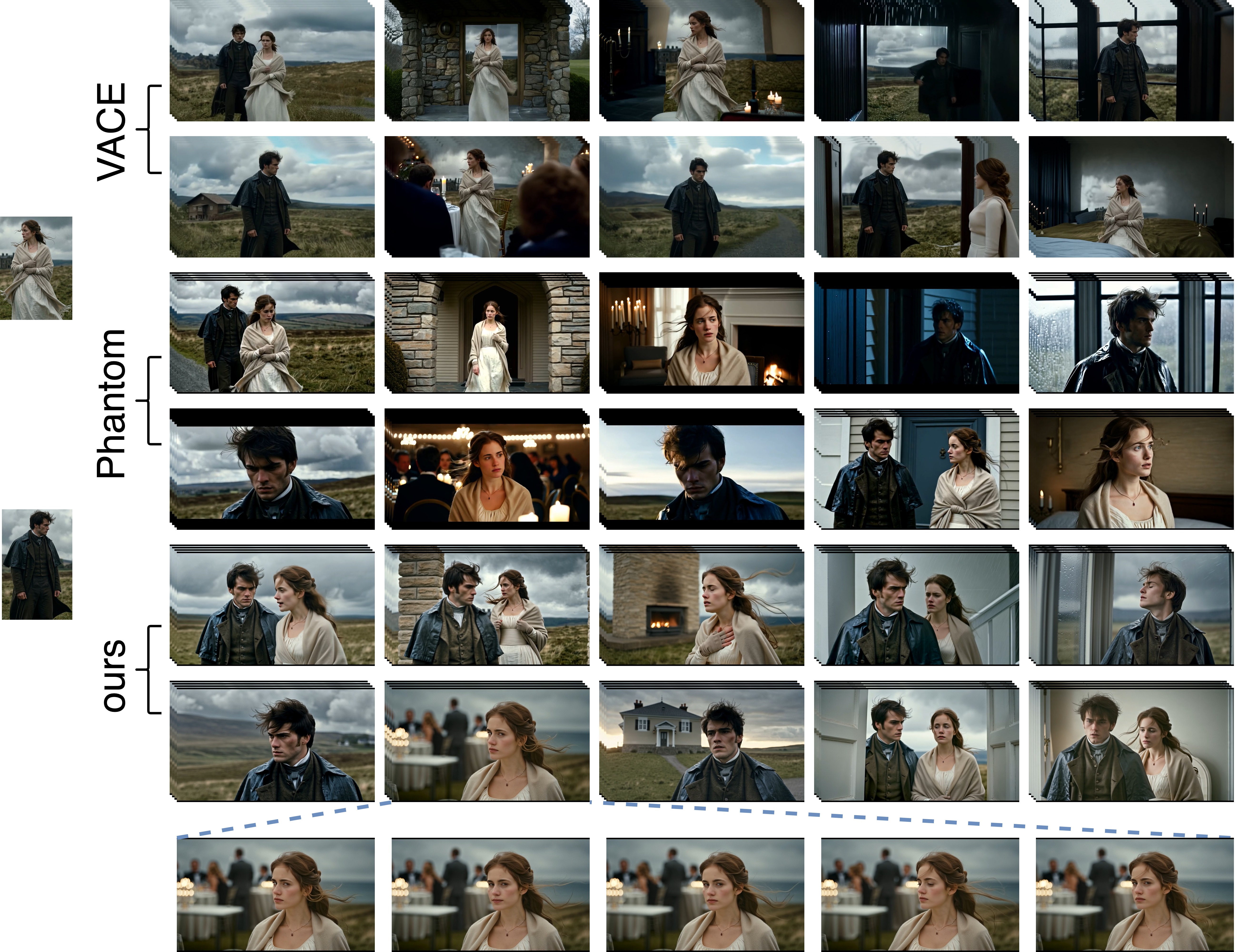}  
\caption{Qualitative Results for Reference-Controllable Multi-Shot T2V.}\label{fig:ref-compare}
\end{figure*}

\section{Experiments}\label{sec:exp}
\subsection{Experimental Settings}
\subsubsection{Implementation Details}
\noindent\textbf{Backbone Models.}
We adopt Wan2.1-14B as the bidirectional DiT-based diffusion backbone for training-free multi-shot text-to-video generation in Section \ref{subsec:multishot_t2v}, and Phantom-14B for reference-controllable multi-shot generation and long-video extension in Sections \ref{subsec:reference_multishot} and \ref{subsec:long_video}. All the parameters of backbones, including the DiT Transformer and VAE encoder/decoder, remain frozen throughout all experiments, and our framework operates purely at inference time without any additional training or fine-tuning. For each video, the number of shots $N_K$, the number of frames per shot $l_k$, and the number of reference images per shot $|\mathcal{R}_k|$ (when applicable) are specified by users and could vary across different experiments. The resolution of generated videos is fixed to 832$\times$480. The number of gap frames $N_G$ is set to 5 for Wan2.1-based multi-shot T2V, and 2 for Phantom-based reference-controllable generation. For long-video extension, we select $N_A = 5$ anchor frames from the first generated shot for jointly producing anchor tokens that remain fixed across subsequent multi-shot segments. All experiments are conducted on NVIDIA H100 GPUs.

\begin{table*}[!t]
\renewcommand{\baselinestretch}{1.0}
\renewcommand{\arraystretch}{1.0}
\setlength{\tabcolsep}{2pt}
\centering
\caption{Component-wise ablation studies on Reference-Controllable Multi-Shot T2V Benchmark.}
\label{tab:ablation_ref_multishot}
\begin{tabular}{@{}l c ccc c c cc@{}}
\toprule
& \multirow{2}*{\textbf{Text Align}$\uparrow$} & \multicolumn{3}{c}{\textbf{Inter-Shot Consistency}$\uparrow$} & \multirow{2}*{\textbf{Transition Deviation}$\downarrow$} & \multirow{2}*{\textbf{Narrative \textbf{Coherence}$\uparrow$}} & \multicolumn{2}{c}{\textbf{Reference Consistency}$\uparrow$} \\
\cmidrule(lr){3-5}\cmidrule(lr){8-9}
 & & \textbf{Semantic} & \textbf{Subject} & \textbf{Scene} & & & \textbf{Background} & \textbf{Subject} \\
\midrule
\rowcolor{gray!12}
\textbf{Proposed} 
& 0.259 & 0.902 & 0.808 & 0.880 & \textbf{6.09} & \textbf{0.699} & \textbf{0.632} & 0.560 \\
w/o Gap Frames 
& 0.259 & 0.913 & 0.831 & 0.901 & 7.12 & 0.658 & 0.629 & 0.537 \\
w/o Transition Frames 
& 0.255 & \textbf{0.930} & 0.822 & 0.903 & 9.55 & 0.673 & 0.594 & 0.514 \\
w/o Ref. Routing 
& 0.253 & 0.877 & 0.778 & 0.818 & 16.96 & 0.583 & 0.629 & \textbf{0.610} \\
w/o Shot-wise VAE Decode 
& \textbf{0.262} & 0.915 & \textbf{0.835} & \textbf{0.915} & 19.14 & 0.677 & 0.631 & 0.540 \\
\bottomrule
\end{tabular}
\end{table*}
\begin{table*}[!t]
\renewcommand{\baselinestretch}{1.0}
\renewcommand{\arraystretch}{1.0}
\setlength{\tabcolsep}{2pt}
\centering
\caption{Ablation studies on the long reference-controllable multi-shot benchmark.}\label{tab:ablation_anchor}
\begin{tabular}{@{}l c ccc c c cc@{}}
\toprule
& \multirow{2}*{\textbf{Text Align}$\uparrow$} & \multicolumn{3}{c}{\textbf{Inter-Shot Consistency}$\uparrow$} & \multirow{2}*{\textbf{Transition Deviation}$\downarrow$} & \multirow{2}*{\textbf{Narrative \textbf{Coherence}$\uparrow$}} & \multicolumn{2}{c}{\textbf{Reference Consistency}$\uparrow$} \\
\cmidrule(lr){3-5}\cmidrule(lr){8-9}
 & & \textbf{Semantic} & \textbf{Subject} & \textbf{Scene} & & & \textbf{Background} & \textbf{Subject} \\
\midrule
\rowcolor{gray!12}
\textbf{Full CineWeaver} $(N_A=5)$
& 0.206 & \textbf{0.556} & \textbf{0.429} & 0.356 & \textbf{2.02} & 0.672 & 0.463 & \textbf{0.476} \\
$N_A=0$ 
&\textbf{0.215}  &0.517  &0.368  &0.272  &2.89  &\textbf{0.709}  &\textbf{0.495} &0.443  \\
$N_A=9$ 
&0.192  &0.537  &\textbf{0.429}  &\textbf{0.401}  &3.54  &0.557  &0.419  &0.444  \\
\bottomrule
\end{tabular}
\end{table*}
\begin{figure*}[!t]
\centering
\includegraphics[width=\linewidth]{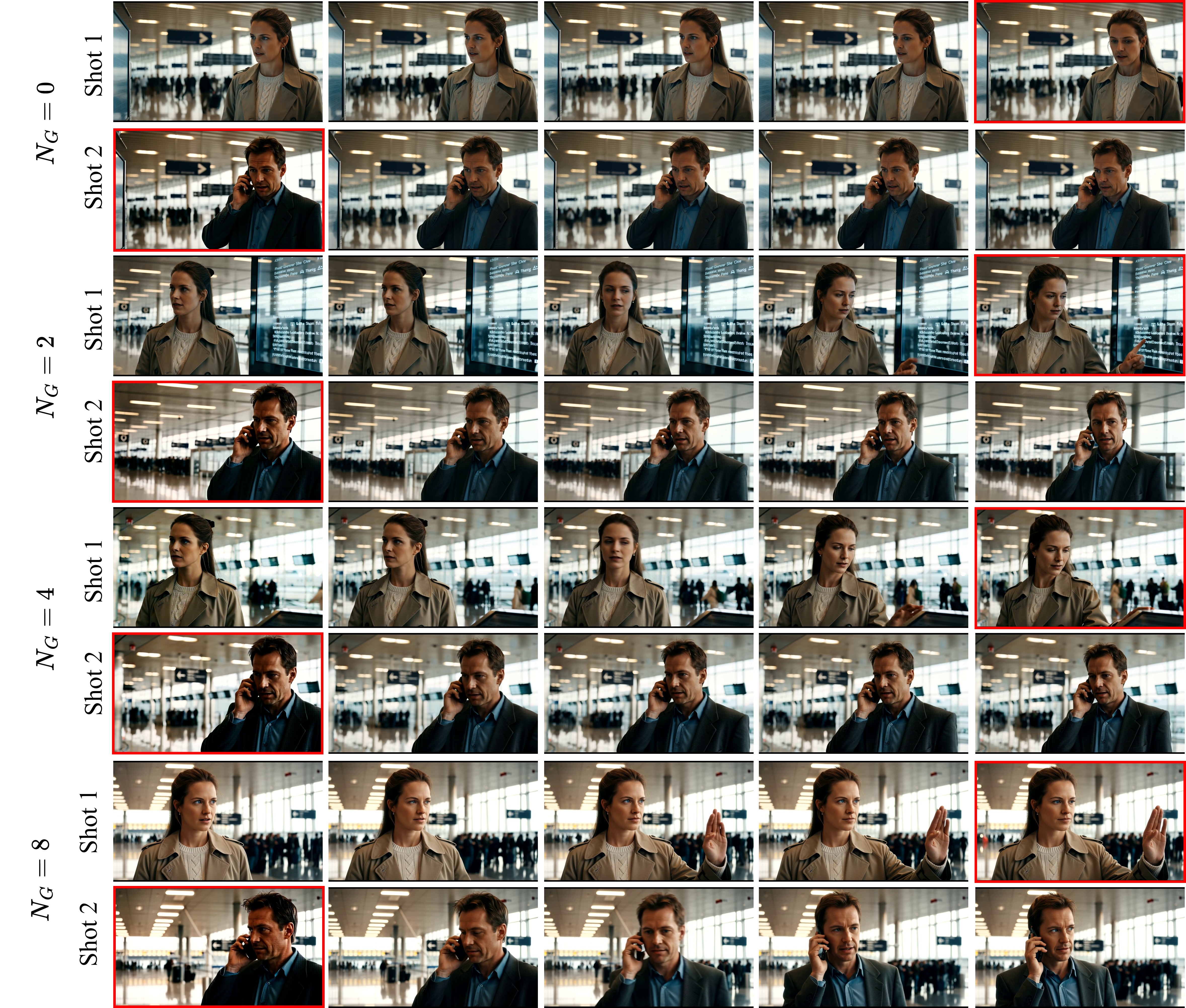}
\caption{Ablation studies on the number of gap frames $N_G$.
Under each setting, the two images highlighted with red boxes indicate the positions where shot transitions occur.}\label{fig:ab-gp}
\end{figure*}

\subsubsection{Comparative Methods}
For multi-shot text-to-video generation, we compare our method with four representative approaches designed for multi-shot video generation, including CineTrans~\citep{cinetrans}, EchoShot~\citep{echoshot}, HoloCine~\citep{holocine}, and MultiShotMaster~\citep{multishotmaster}, which explicitly model shot transitions and structured multi-shot generation under different architectural or training paradigms. 

For reference-controllable multi-shot generation, MultiShotMaster~\citep{multishotmaster} is the only prior work that supports both multi-shot generation and reference guidance within a unified framework. However, its publicly released implementation provides only the text-to-video component, while the reference-guided module is not available. For fair comparison under the reference-controllable setting, we follow the experimental protocol adopted by MultiShotMaster \citep{multishotmaster}. We first generate each shot independently using strong reference-guided single-shot video generation models Phantom~\citep{phantom} and VACE~\citep{vace}, and then concatenate the generated shots to obtain multi-shot videos. This reproduces a competitive reference-controllable multi-shot baseline in practice.

\subsubsection{Benchmark}
We construct three evaluation benchmarks to assess multi-shot generation, reference controllability, and long-form consistency. i) \textbf{Multi-Shot T2V Benchmark.} Following MultiShotMaster~\citep{multishotmaster}, we generate 200 multi-shot prompt sets using Gemini-3 \citep{gemini3}. Each video consists of 2–4 shots, with the length of each shot specified by Gemini. Prompts are randomly sampled to cover diverse scenes, actions, and transitions. ii) \textbf{Reference-Controllable Multi-Shot T2V Benchmark.} We also follow MultiShotMaster~\citep{multishotmaster} to evaluate three reference injection modes: background injection, subject injection, and joint background+subject injection. For each mode, we generate 50 multi-shot prompt sets using Gemini-3 \citep{gemini3}. Gemini additionally specifies the required reference images for each shot, which are synthesized using Nano Banana~\citep{nanobanana}. Each video consists of 2–3 shots, with shot lengths determined by Gemini. iii) \textbf{Long Reference-Controllable Multi-Shot T2V Benchmark.} To evaluate long-form generation, we construct 15 prompt sets using Gemini-3~\citep{gemini3}, each corresponding to a long video containing more than 30 shots and exceeding one minute in duration. Each prompt set follows a consistent global theme. For each shot, subject and background reference images are specified and generated using Nano Banana~\citep{nanobanana}. This benchmark design allows us to separately evaluate the three key capabilities required by cinematic video generation. The first benchmark focuses the ability of explicit shot transitions from shot-level prompts. The second benchmark evaluates the correctness in injecting reference images into designated shots without disrupting multi-shot generation. The third benchmark stresses long-range consistency and requires models to maintain coherent appearance across dozens of shots.

\subsubsection{Metrics}
Following MultishotMaster~\citep{multishotmaster}, we evaluate generation quality from five perspectives, including text alignment, inter-shot consistency, transition accuracy, narrative coherence, and reference injection consistency. All metrics are computed using the same pretrained models and a unified evaluation pipeline to ensure fair comparison, as elaborated below.

\textbf{Text Alignment.} We compute video--text cosine similarity using ViCLIP~\citep{viclip} between each generated shot and its corresponding prompt. For each shot, we uniformly sample 8 frames and resize them to 224$\times$224 before feature extraction. Video and text embeddings are obtained using the pretrained ViCLIP model, and cosine similarity is computed between them. Shot-level similarities are then averaged to obtain the final video-level alignment score.

\textbf{Inter-Shot Consistency.} We evaluate cross-shot stability from three aspects: semantic, subject, and scene consistency. Semantic consistency is measured using ViCLIP \citep{viclip} video embeddings computed from sampled frames of each shot. Subject and scene consistency are evaluated using object-aware features. Specifically, we first detect foreground objects using a pretrained YOLOv11 \citep{yolov11} detector with a confidence threshold of 0.3 and refine object masks with SAM \citep{sam} to obtain precise subject regions. Subject crops and background regions are then extracted accordingly. DINOv2 \citep{dinov2} features (ViT-L/14) are computed for both subject crops and background regions, and cosine similarity between adjacent shots is used to quantify cross-shot consistency.

\textbf{Transition Deviation.} We detect shot boundaries using a pretrained TransNetV2~\citep{transnetv2} model. Following the standard TransNetV2 preprocessing, frames are resized to 48$\times$27 before inference. Transition probabilities are predicted using a sliding window of 100 frames with a stride of 50, and transitions are determined with a threshold of 0.5. We then compute the mean frame deviation between predicted transitions and ground-truth transitions derived from metadata using Hungarian matching.

\textbf{Narrative Coherence.} We employ Gemini-2.5~\citep{gemini2.5} to evaluate scene consistency, subject consistency, action coherence, and spatial consistency. For each video, we proportionally sample 20 frames across shots while ensuring at least one frame per shot. The sampled frames, together with shot-level and global descriptions, are provided to the model as structured inputs. Gemini returns binary judgments for each dimension, and the final score is computed by averaging the four dimensions.

\begin{figure*}[!t]
\centering
\includegraphics[width=\linewidth]{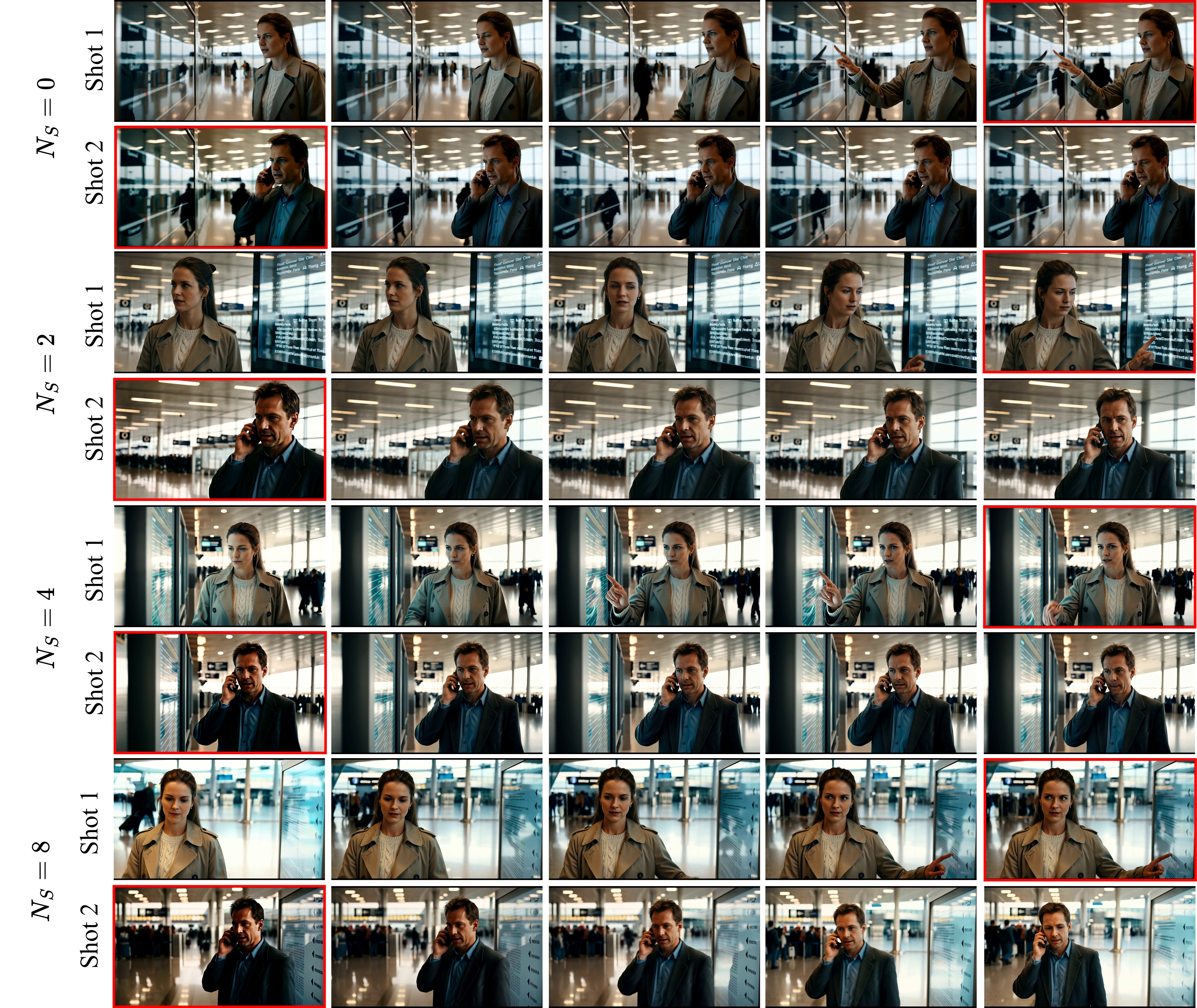}
\caption{Ablation studies on the number of transition frames $N_S$. Under each setting, the two images highlighted with red boxes indicate the positions where shot transitions occur.}
\label{fig:ab-ns}
\end{figure*}

\textbf{Reference Injection Consistency.} For the task of reference-controllable generation, we employ YOLOv11~\citep{yolov11} and SAM~\citep{sam} to extract subject and background regions from both reference images and generated frames. For each shot, keyframes are uniformly sampled and processed to obtain subject crops and background regions. DINOv2~\citep{dinov2} embeddings are extracted from these regions using the CLS token representation, and cosine similarity between reference and generated features is used to measure subject-level and background-level preservation. Frame-level similarities are first aggregated within each shot and then averaged across shots to obtain the final consistency score.

\begin{figure*}[t]
\centering
\includegraphics[width=\linewidth]{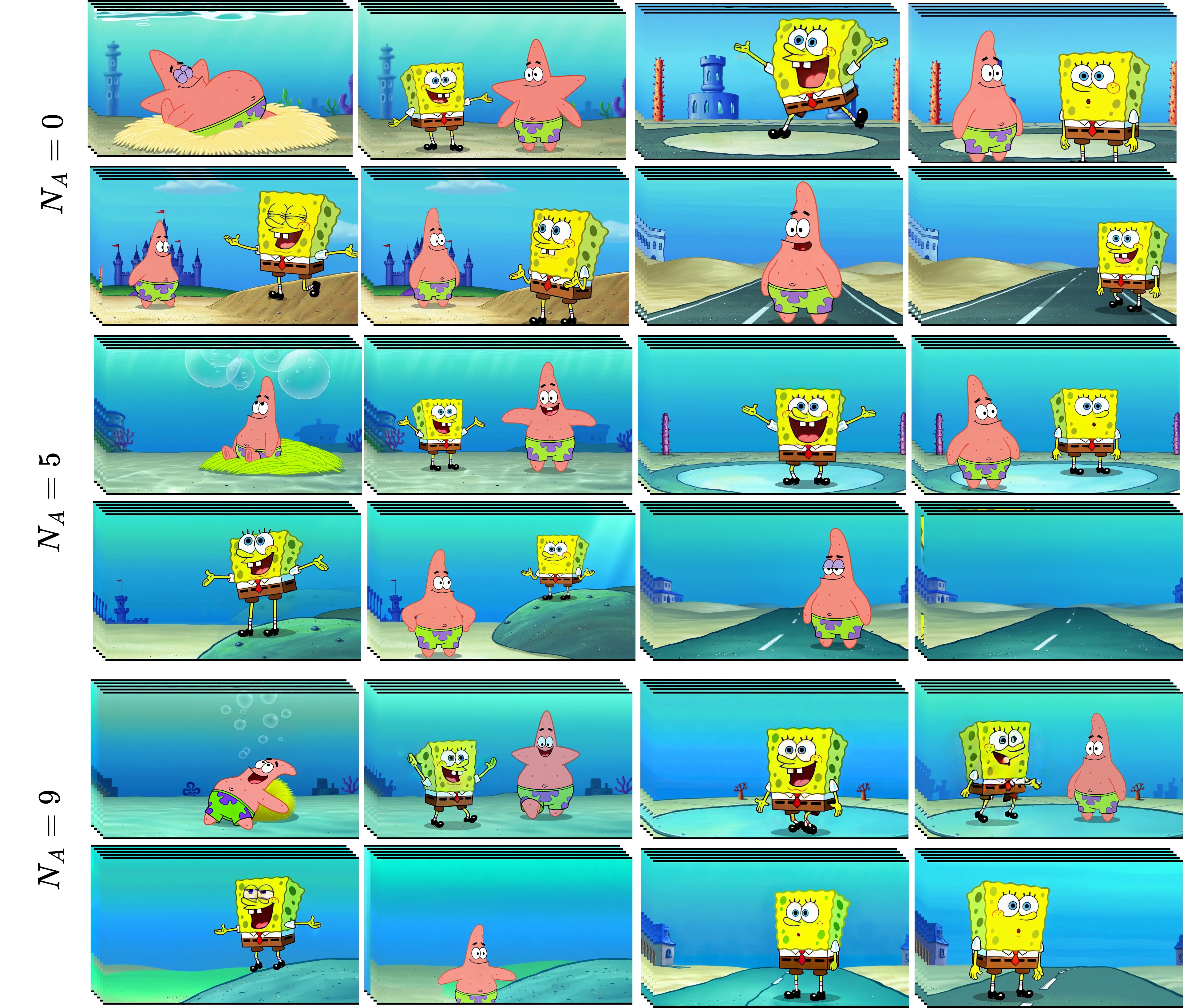}
\caption{Ablation studies on the number of anchor frames $N_A$.}\label{fig:ab-anchor}
\end{figure*}

\subsection{Quantitative Results}
Table~\ref{tab1:main results} reports quantitative results across the three settings. For \textbf{Multi-shot Text-to-Video}, CineWeaver achieves the best overall performance, with the lowest transition deviation and the highest inter-shot consistency and narrative coherence. This indicates that our training-free design can produce clear shot boundaries while preserving cross-shot coherence.

For \textbf{Reference-Controllable Multi-shot Text-to-Video}, CineWeaver achieves the strongest inter-shot semantic and subject consistency, the lowest transition deviation, and substantially higher narrative coherence than independently concatenating Phantom~\citep{phantom} or VACE~\citep{vace}. Although VACE obtains higher background reference similarity, we provide a better balance between reference controllability, shot transition accuracy, and multi-shot coherence. 
For \textbf{Long Reference-Controllable Multi-shot Text-to-Video}, CineWeaver significantly improves transition accuracy, narrative coherence, and inter-shot consistency over the baselines. These results show that anchor memory mitigates long-form drift and maintains a coherent global appearance across extended multi-shot videos.

\subsection{Qualitative Results}
We first present the qualitative results for \textbf{multi-shot T2V generation}. \figurename~\ref{fig:t2v-compare} shows that the proposed \textit{training-free} CineWeaver produces videos with clear shot transitions and strong visual coherence and yields comparable results to \textit{training-based} multi-shot T2V approaches~\citep{echoshot,holocine,cinetrans,multishotmaster} without additional training. 

Furthermore, we present the qualitative results for \textbf{long-term reference-controllable multi-shot T2V generation} in \figurename~\ref{fig:ref-compare}. VACE~\citep{vace} cannot preserve identity consistency between generated shots given reference images of characters. In several shots, the character’s pose closely replicates that of the reference image, indicating limited controllability in generating diverse yet identity-consistent motions. Phantom~\citep{phantom} is noticeably weaker in global consistency across shots, and causes significant variations in lighting, tone, and overall appearance between different shots. In contrast, CineWeaver produces clear shot transitions and faithfully follows the reference images. It maintains stronger identity consistency and more stable global appearance across shots. 

\subsection{Ablation Studies}
\subsubsection{Quantitative Evaluations}

Quantitative ablation studies are conducted on both the reference-controllable and long reference-controllable multi-shot benchmarks to evaluate the contribution of each proposed component. Specifically, we first investigate the roles of gap frames, transition frames, shot-routed reference conditioning, and isolated VAE decoding on the reference-controllable benchmark. Besides, we further investigate the roles of Anchor Memory on the long reference-controllable multi-shot benchmark.

Table~\ref{tab:ablation_ref_multishot} shows that removing each component causes performance loss. Without gap frames or transition frames, the transition deviation increases, indicating less accurate shot boundaries. Removing reference routing causes the largest transition error and a clear drop in narrative coherence, showing that shot-routed reference conditioning is important to prevent cross-shot reference interference. Joint VAE decoding also significantly worsens transition accuracy, confirming the necessity to reset the decoding state to avoid cache-induced visual leakage. Although some variants achieve higher inter-shot similarity, this is partly because weaker shot separation makes adjacent shots more visually similar and does not necessarily indicate better multi-shot generation.

Table~\ref{tab:ablation_anchor} further studies the effect of Anchor Memory in long-form generation. Without anchor frames ($N_A=0$), the model suffers from weaker inter-shot consistency and transition accuracy. Increasing $N_A$ to 9 improves scene consistency but affects transition accuracy and narrative coherence. Thus, we adopt $N_A=5$ as the default setting to better balance among transition accuracy, long-form consistency, and reference preservation.

\subsubsection{Qualitative Evaluations}
We perform qualitative ablation studies to analyze the impact of three key components with varying parameters, including the number of gap frames $N_G$, the number of transition frames $N_S$, and the number of anchor frames $N_A$. In particular, $N_G=0$ corresponds to removing gap frames, $N_S=0$ corresponds to removing transition frames, and $N_A=0$ corresponds to disabling the anchor memory mechanism. Since CineWeaver can generate both long cinematic videos and long animated sequences, we randomly select several examples to conduct the ablation studies.

\textbf{Number of Gap Frames $N_G$.}
We generate two consecutive shots with different values of $N_G$ while keeping all other settings unchanged to better illustrate the effect of $N_G$ on shot transition. 
We evaluate several settings, including $N_G=0$, $N_G=2$, $N_G=4$, and $N_G=8$, and the results are shown in \figurename~\ref{fig:ab-gp}. Under each setting, the two frames highlighted with red boxes indicate the positions where shot transitions occur. When $N_G=0$, i.e., without gap frames, the model's ability to perform shot transitions is weakened, demonstrating the necessity of introducing gap frames. 
When $N_G=2$, the model is already able to produce clear and natural shot transitions. 
Further increasing $N_G$ leads to higher computational costs during inference. Therefore, $N_G$ is set to 2 by default for reference-controllable generation.

\textbf{Number of Transition Frames $N_S$.}
Similarly, $N_S$ mainly affects the quality of shot transitions. Therefore, we also generate two consecutive shots with different values of $N_S$ while keeping all other settings unchanged. 
By default, for reference-controllable generation, we set $N_S=2$. 
We evaluate several settings, including $N_S=0$, $N_S=2$, $N_S=4$, and $N_S=8$, and the results are shown in \figurename~\ref{fig:ab-ns}.
Similar conclusions can be drawn for $N_S$. When $N_S=0$, the ability to produce shot transitions is weakened, while increasing $N_S$ further results in higher computational costs. Therefore, we adopt $N_S=2$ as the default setting.

\textbf{Number of Anchor Frames $N_A$.}
To analyze the effect of the number of anchor frames $N_A$, for each setting, we generate 4 groups of results, where each group consists of 2 consecutive shots, resulting in 8 shots in total. 
We evaluate several settings with $N_A=0$, $N_A=5$, and $N_A=9$. 
Due to the temporal compression of the Wan VAE, $N_A$ must follow the form $4n+1$. Here, $N_A=0$ denotes the case without using anchor frames. For $N_A=5$ and $N_A=9$, we use the same anchor shot. The difference lies in the number of frames sampled from the anchor shot: the first 5 frames are used as anchor frames when $N_A=5$, while the first 9 frames are used when $N_A=9$. As shown in \figurename~\ref{fig:ab-anchor}, compared with the setting without anchor frames, introducing anchor frames improves the consistency of global appearance, such as color tone and visual style across shots.  According to our observations, $N_A=5$ provides a good trade-off. Further increasing $N_A$ leads to a slight degradation in video quality while also increasing the computational cost.

\begin{table}[!t]
\renewcommand{\baselinestretch}{1.0}
\renewcommand{\arraystretch}{1.0}
\setlength{\abovecaptionskip}{0pt}
\centering
\caption{User study results of the preferences ratio for Cineweaver.}\label{tab:user}
\begin{tabular}{@{}lcccc@{}}
\toprule
Method
& \makecell{Transition\\Clarity} & \makecell{Reference\\Consistency} & \makecell{Shot\\Coherence} & \makecell{Narrative\\Coherence}\\
\midrule
vs. Phantom &  \textbf{61.2}  &  \textbf{51.3}  &  \textbf{76.8}  &  \textbf{83.4}   \\
vs. VACE &  \textbf{58.4}  &  49.6  &  \textbf{71.5}  &  \textbf{77.6}    \\
\bottomrule
\end{tabular}
\end{table}

\subsection{User Study}
A user study is made to complement automatic metrics. We randomly sample video pairs from the reference-controllable multi-shot and long-form benchmarks. Each pair contains two anonymized videos generated from the same shot-level prompts and reference images, with the display order randomly shuffled. Participants compare the videos by choosing A, B, or tie under five criteria (\emph{i.e.}, transition clarity, reference consistency, shot coherence, narrative coherence, and overall quality). Ties are counted as half votes for A and B. For each criterion, the preference score of method A is computed as
\[
\mathrm{Pref}(A)=
\frac{N_A + 0.5N_{\mathrm{tie}}}
{N_A + N_B + N_{\mathrm{tie}}},
\]
where $N_A$, $N_B$, and $N_{\mathrm{tie}}$ denote the numbers of votes for A, B, and tie, respectively. Table~\ref{tab:user} reports the preference percentage for CineWeaver. Participants generally prefer CineWeaver over Phantom and VACE, especially in shot coherence and narrative coherence.

\begin{figure*}[!t]
\renewcommand{\baselinestretch}{1.0} 
\setlength{\abovecaptionskip}{0pt}
\centering
\includegraphics[width=\linewidth]{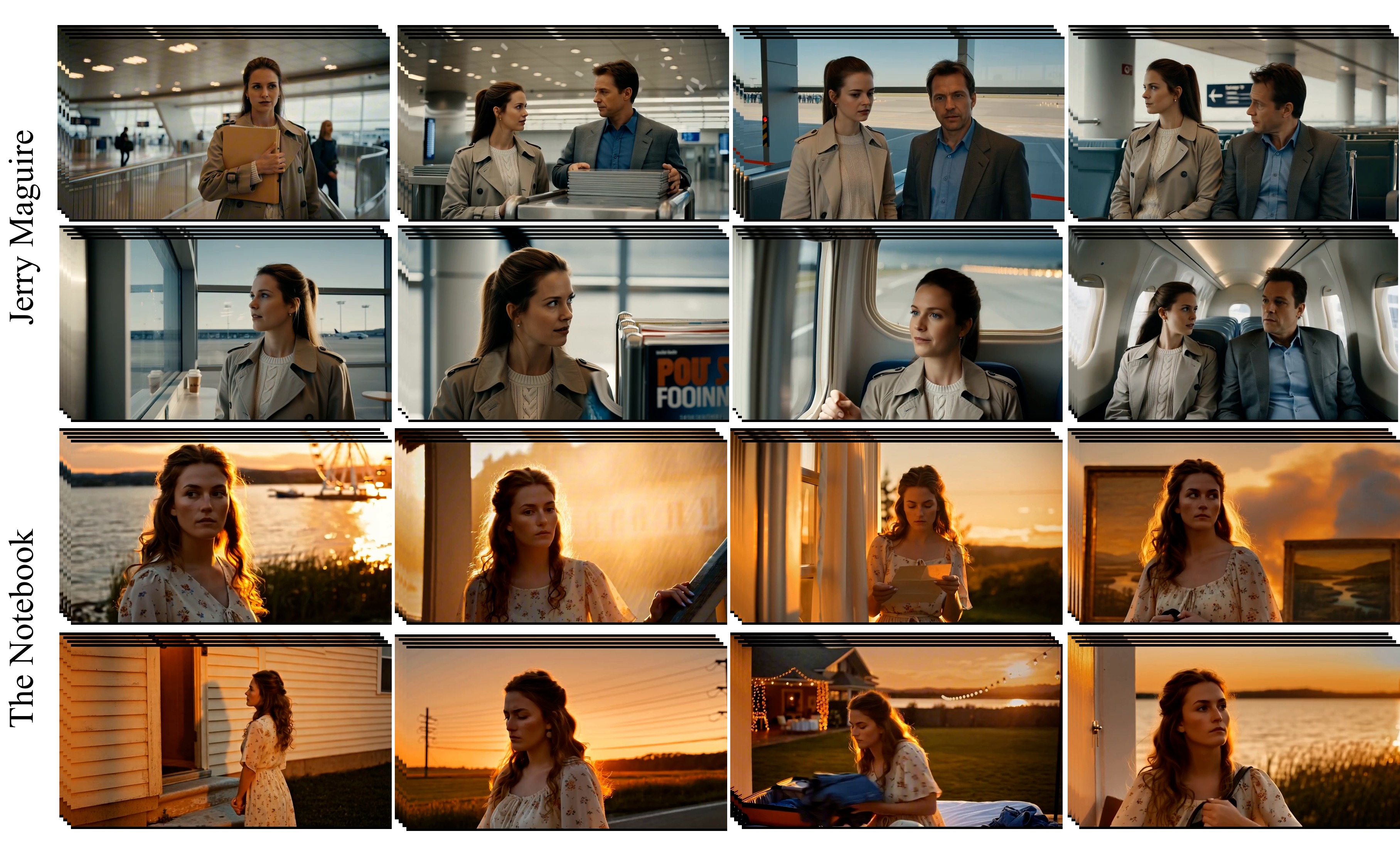}  
\caption{Cinematic multi-shot videos generated by CineWeaver.}\label{fig:more-results-cine-pick}
\end{figure*}
\begin{figure*}[!t]
\renewcommand{\baselinestretch}{1.0} 
\setlength{\abovecaptionskip}{0pt}
\centering
\includegraphics[width=\linewidth]{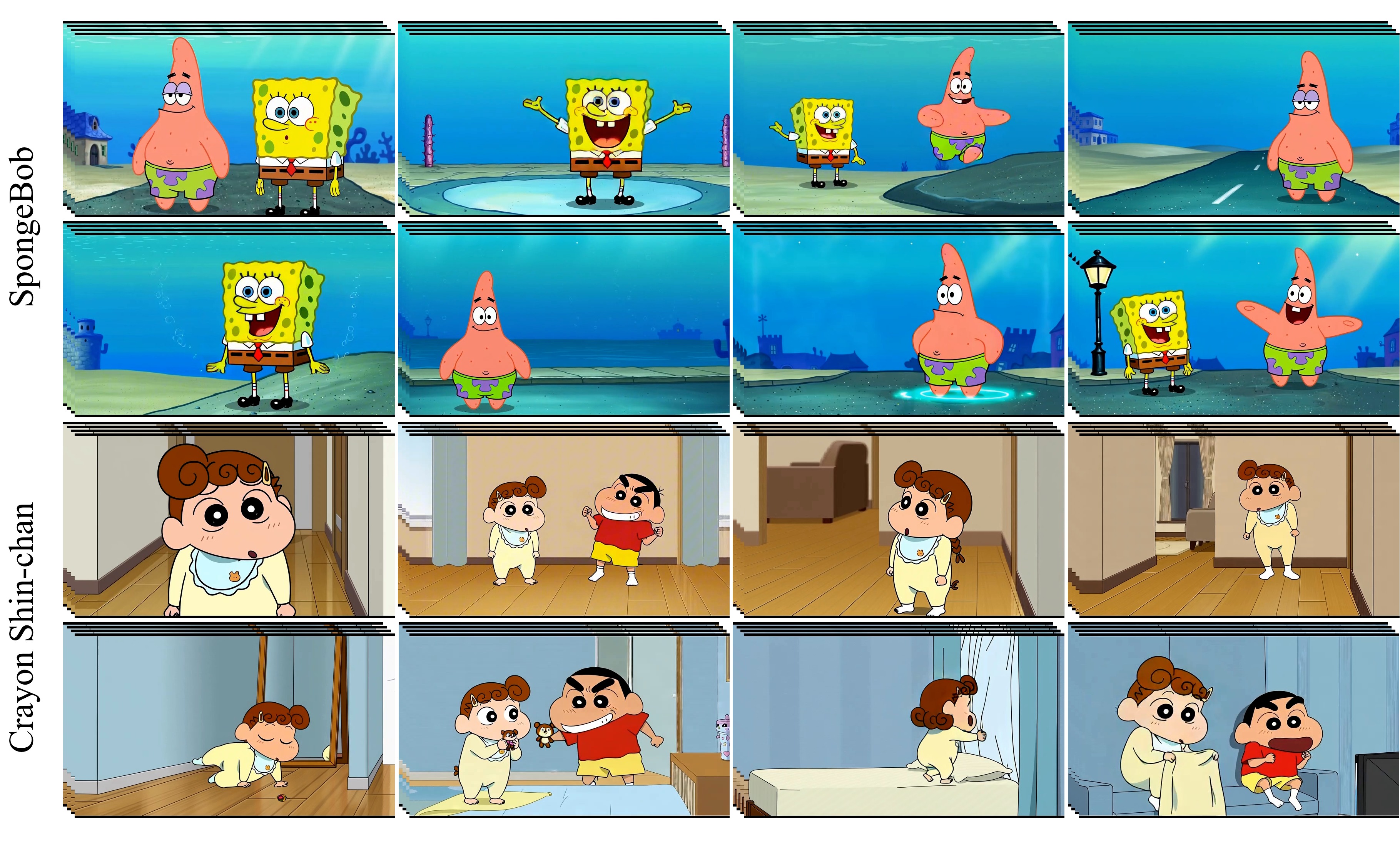}  
\caption{Animated multi-shot videos generated by CineWeaver.}\label{fig:more-results}
\end{figure*}
\begin{figure*}[!t]
\centering
\includegraphics[width=\linewidth]{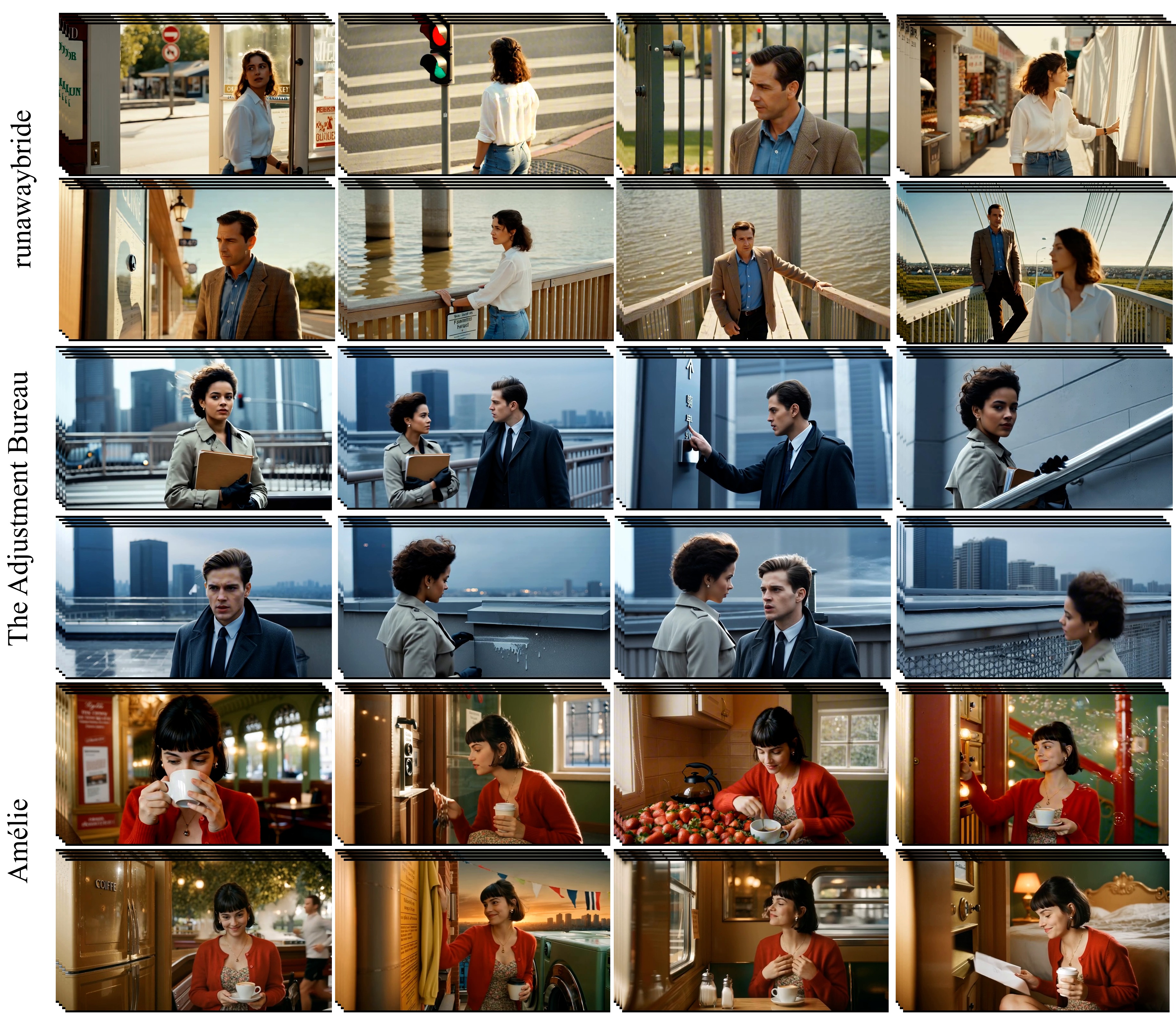}
\caption{Supplemental examples for long cinematic videos generated by CineWeaver.}\label{fig:long-cine}
\end{figure*}
\begin{figure*}[!t]
\centering
\includegraphics[width=\linewidth]{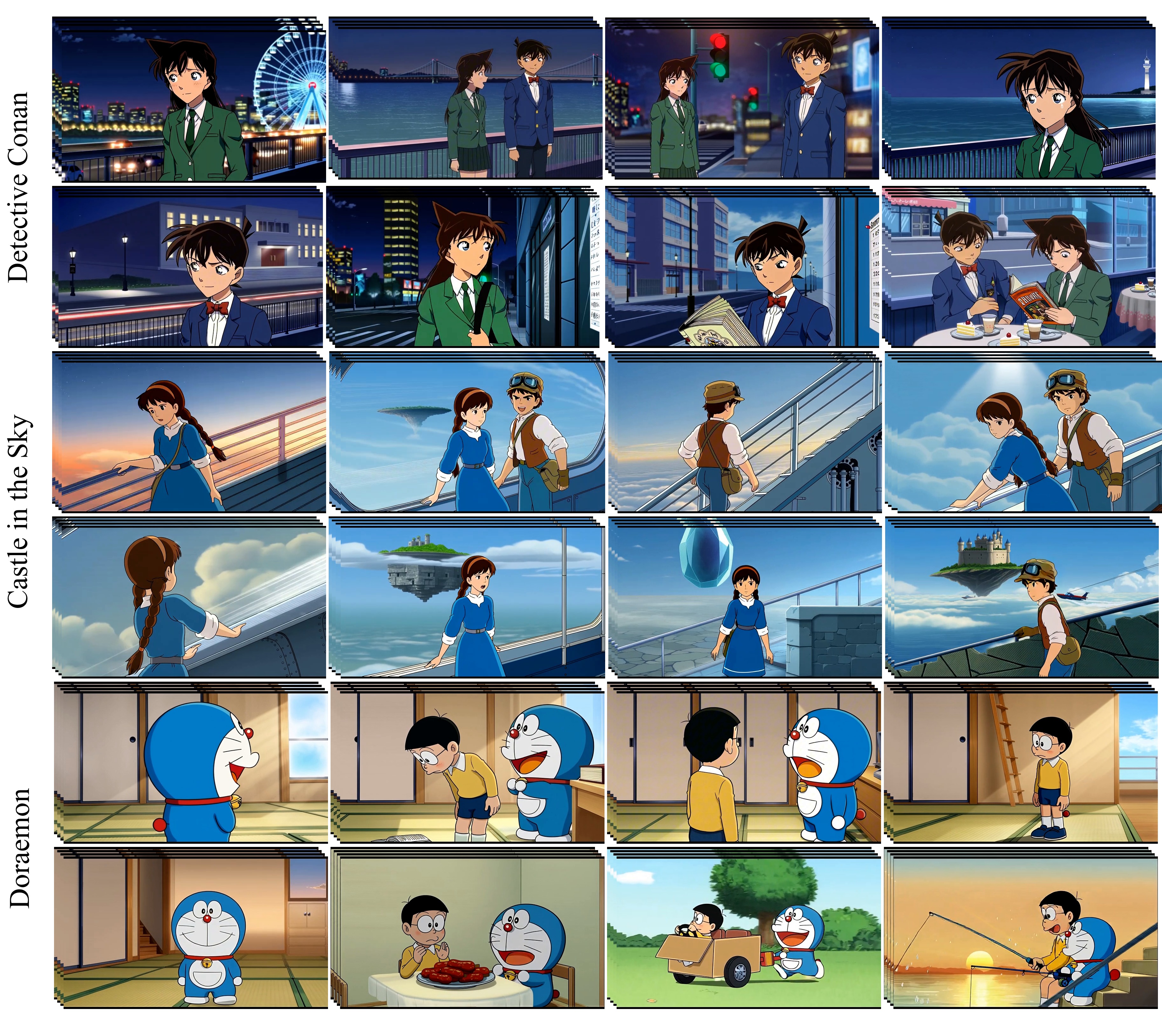}
\caption{Supplemental examples for long animated videos generated by CineWeaver.}
\label{fig:long-animated}
\end{figure*}

\subsection{Versatility Analysis}
We further demonstrate the versatility of CineWeaver across different multimedia generation scenarios. 
First, we show more \textbf{cinematic multi-shot videos} in Fig.~\ref{fig:more-results-cine-pick}. These examples are inspired by the narratives of \emph{The Notebook} and \emph{Jerry Maguire}, while the character reference images are synthesized by Nano Banana. CineWeaver generates coherent cinematic multi-shot videos from shot-level prompts and reference images, while maintaining clear shot transitions and consistent character appearance. 
Second, Fig. \ref{fig:more-results} shows \textbf{animated multi-shot videos} based on \emph{SpongeBob SquarePants} and \emph{Crayon Shin-chan} directly using the original characters as references for generating animated multi-shot stories. 
CineWeaver generalizes well to stylized animation domains and preserves reference-consistent appearance and coherent global visual style. 

CineWeaver is not tied to a specific reference source or visual domain. For example, references are synthesized portraits that provide realistic character appearances in cinematic examples and are original cartoon characters with highly stylized shapes and colors in animated examples. 
In both cases, the same inference framework can route references to designated shots and maintain coherent multi-shot generation. It is flexible across various scenarios. Refer to the supplementary material for full video demos.

\subsection{Supplemental Experimental Results}
We provide additional results of long-form reference-controllable multi-shot videos generated by CineWeaver. To demonstrate the versatility of our method, we generate two types of content, \emph{i.e.}, long cinematic videos and long animated videos. 
For the cinematic videos, instead of using screenshots of the original movie characters as references, we \textbf{generate reference characters} using Nano-Banana and synthesize videos inspired by several films, including \textit{Runaway Bride}, \textit{The Adjustment Bureau}, and \textit{Amélie}. 
For the animated videos, we \textbf{directly use the original characters as references} rather than generating new ones, and produce videos based on \textit{Doraemon}, \textit{Detective Conan}, and \textit{Castle in the Sky}. We also provide all reference images.

Most of the generated videos reach minute-level durations and contain dozens of shots. 
Due to space limitations in the PDF, we show only 8 shots for each example. We show the examples of long cinematic videos in \figurename~\ref{fig:long-cine} and long animated videos in \figurename~\ref{fig:long-animated}. We also include the full videos (in \texttt{.mp4} format) containing more shots as multimedia supplementary material.


\section{Conclusion}\label{sec:con}
In this paper, we present CineWeaver, a training-free framework for reference-controllable multi-shot long video generation. Our key insight is that multi-shot composition does not require learning a new capability, but can be achieved by explicitly manipulating temporal continuity during inference. Based on this insight, by breaking the inherent temporal continuity bias of pretrained video diffusion models during inference, CineWeaver enables multi-shot composition without retraining. Building upon this training-free structure, we introduce shot-wise reference routing for fine-grained controllability and Anchor Memory for maintaining global appearance consistency across independently generated segments. Extensive experiments demonstrate that CineWeaver successfully provides a unified framework for cinematic video generation with clear shot transitions, strong reference adherence, and long-form coherence.

\section*{Acknowledgments}
This work was supported in part by the National Natural Science Foundation of China under Grant 62431017.

\section*{Data availability}
Data will be made available on reasonable request.

\bibliographystyle{spbasic}
\bibliography{references}

\vfill

\end{document}